\documentclass[10pt,twocolumn,letterpaper]{article}

\usepackage{wacv}
\usepackage{times}
\usepackage{epsfig}
\usepackage{graphicx}
\usepackage{amsmath}
\usepackage{amssymb}

\usepackage{multirow, dcolumn}
\usepackage{subcaption}
\usepackage{lipsum}
\usepackage{footnote}
\usepackage{siunitx}
\usepackage{makecell}
\usepackage{epstopdf}
\usepackage{fancyhdr}
\usepackage{xcolor}
\usepackage{gensymb}
\usepackage[normalem]{ulem}
\usepackage{soul}
\usepackage{verbatim}
\usepackage{booktabs}
\usepackage{pifont}
\usepackage{tablefootnote}
\newcommand{\cmark}{\ding{51}}%
\newcommand{\xmark}{\ding{55}}%

\usepackage{adjustbox}

\wacvfinalcopy 

\ifwacvfinal
\def\assignedStartPage{1} 
\fi

\ifwacvfinal
\usepackage[breaklinks=true,bookmarks=false]{hyperref}
\else
\usepackage[pagebackref=true,breaklinks=true,colorlinks,bookmarks=false]{hyperref}
\fi

\ifwacvfinal
\else
\fi

\begin{document}

\title{MM-ViT: Multi-Modal Video Transformer for \\ Compressed Video Action Recognition}

\author{Jiawei Chen\\
OPPO US Research Center\\
{\tt\small jiawei.chen@oppo.com}
\and
Chiu Man Ho \\
OPPO US Research Center\\
{\tt\small chiuman@oppo.com}
}

\maketitle

\begin{abstract}
This paper presents a pure transformer-based approach, dubbed the Multi-Modal Video Transformer (MM-ViT), for video action recognition. 
Different from other schemes which solely utilize the decoded RGB frames, MM-ViT operates exclusively in the compressed video domain and exploits all readily available modalities, i.e., I-frames, motion vectors, residuals and audio waveform. 
In order to handle the large number of spatiotemporal tokens extracted from multiple modalities, we develop several scalable model variants which factorize self-attention across the space, time and modality dimensions.
In addition, to further explore the rich inter-modal interactions and their effects,
we develop and compare three distinct cross-modal attention mechanisms that can be seamlessly integrated into the transformer building block. 
Extensive experiments on three public action recognition benchmarks (UCF-101, Something-Something-v2, Kinetics-600) demonstrate that MM-ViT outperforms the state-of-the-art video transformers in both efficiency and accuracy, and performs better or equally well to the state-of-the-art CNN counterparts with computationally-heavy optical flow. 

\end{abstract}

\section{Introduction}
%
%
%
%

%

Video is  one of the most popular media forms due to its rich visual and auditory content.
%
In particular, videos have accounted for 75$\%$ of the global IP traffic everyday~\cite{pepper2013cisco}, which in consequence, poses an urgent need for automated video understanding methods. 
%
%
%
Among the video analysis tasks, action recognition is a fundamental one and becomes increasingly demanding in video applications, e.g., intelligent surveillance, self-driving, personal recommendation and entertainment~\cite{poquet2018video}. 
However, most existing action recognition methods do not fully exploit the rich multi-modal information within videos, as they either depend on a single modality~\cite{arnab2021vivit,bertasius2021space,feichtenhofer2019slowfast} or treat modalities separately~\cite{chen2017semi,shou2019dmc,simonyan2014two,wu2018compressed}.
%
%
This is partially due to the difficulty of reasoning expressive cross-modal interactions~\cite{hendricks2021decoupling}. 
Whereas, recent advancements in cross-modal vision-language transformers~\cite{gabeur2020multi,radford2021learning,tan2019lxmert} and multi-modal multi-task transformers~\cite{hu2021transformer} amply demonstrate the superiority of transformers in reasoning across multiple modalities.
Furthermore, pure transformer models have also achieved competitive performance for vision tasks, e.g., image classification~\cite{dosovitskiy2020image,liu2021swin}, object detection~\cite{liu2021swin,zhu2020deformable} and video classification~\cite{arnab2021vivit,bertasius2021space}.
Such successes motivate our attempt to design a new transformer-based approach which explicitly interprets both visual and audio concepts for action recognition. 
%
\begin{figure}
\begin{center} 
\includegraphics[width=1.0\linewidth]{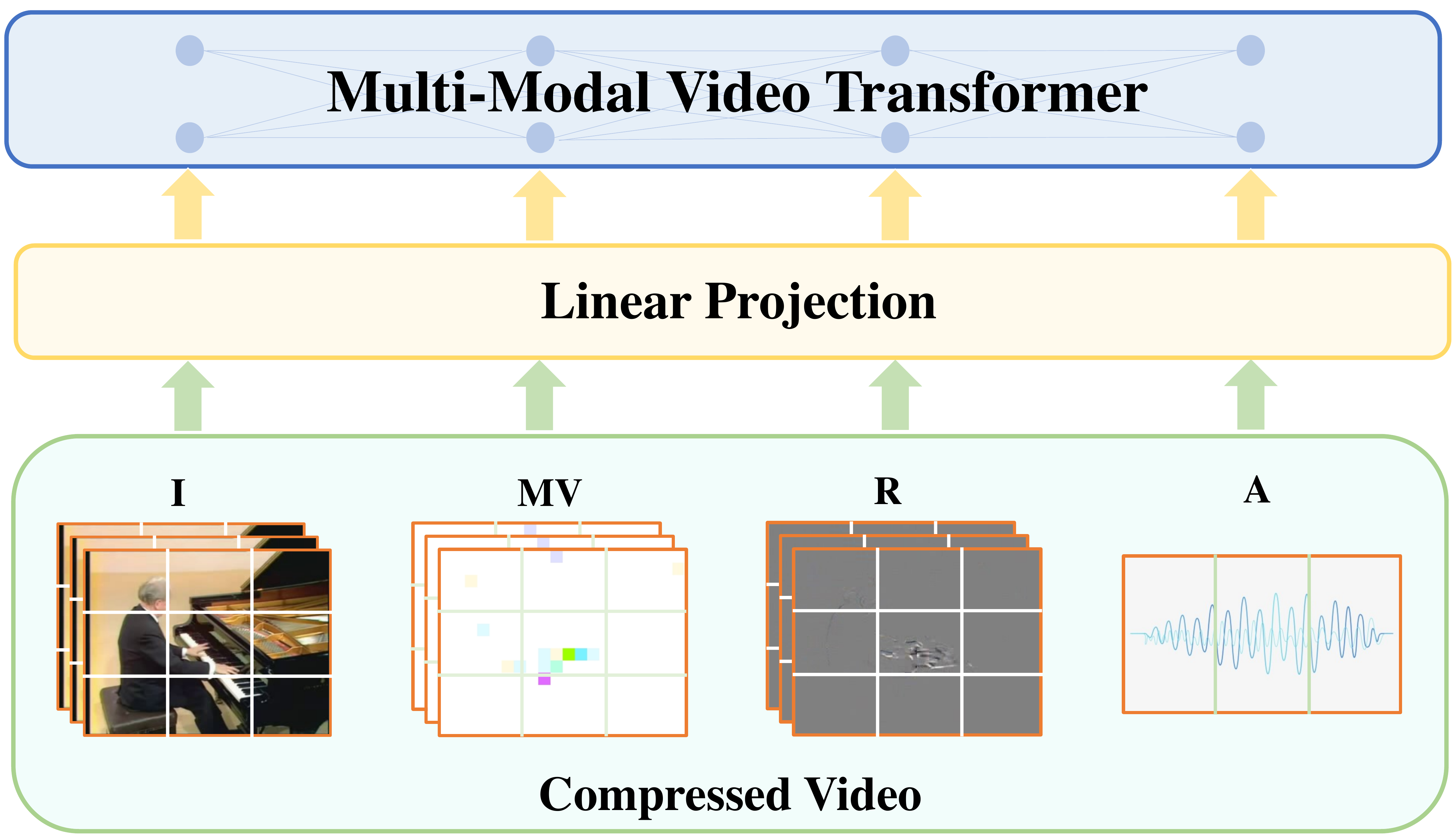}
\end{center}
\vglue -0.4cm
\caption{Our proposed MM-ViT operates exclusively in the compressed video domain, which allows collaborative fusion of appearance (I-frame), motion (Motion-vector $\&$ Residual) and audio features for action recognition. To the best of our knowledge, MM-ViT serves as the first work combining the strength of Transformer and multi-modal learning for compressed video action recognition. }
\label{Fig:preview}
\vglue -0.4cm
\end{figure}

Another persistent challenge in action recognition is to effectively and efficiently model the temporal structure with large variations and complexities.
Early works directly apply 2D CNNs to individual frames and then aggregate frame-level features via mid-level fusion or late-fusion~\cite{karpathy2014large,wang2016temporal,zhou2018temporal}. 
Later, 3D/Pseudo-3D CNNs~\cite{carreira2017quo,chen2020residual,ji20123d,qiu2017learning,tran2015learning,tran2018closer} propose to operate convolution jointly over space and time to better integrate temporal information with CNNs. 
Recently, video transformers equipped with spatial-temporal self-attention are proposed for video classification and show promising results~\cite{arnab2021vivit,bertasius2021space}.
Nevertheless, most state-of-the-art results~\cite{fan2019more,lin2019tsm,stroud2020d3d,tran2018closer,xie2018rethinking} on public benchmarks are still achieved by incorporating hand-crafted motion features, i.e., dense optical flow. 
It indicates that the optical flow extracted by external off-the-shelf methods contains complementary motion cues to the spatiotemporal features produced by deep learning frameworks. 
However, obtaining optical flow frame-by-frame is excessively time consuming, which usually causes a major computational bottleneck in recognition models. 
A recent line of works~\cite{shou2019dmc,wu2018compressed,zhang2016real,zhang2018real} avoid optical flow computation by exploiting the motion information from compressed videos. 
Such a video contains only a few key frames (i.e., \textit{I-frames}) and their offsets
(i.e., \textit{motion vectors} and \textit{residuals}) for storage reduction.
Specifically, these methods utilize the \textit{motion vectors} and \textit{residuals} present in the compressed video to model motion, and manage to achieve competitive accuracy, while run at least an order of magnitude faster than competing methods with optical flow. 
%
%
%
Despite being promising, these methods fail to take advantage of the strong interaction between modalities (i.e., treating modalities separately), and the crucial information in the audio stream.
%

To address the aforementioned issues in the existing methods, we propose a pure-transformer approach, named Multi-Modal Video Transformer (MM-ViT), for video action recognition.
The design of MM-ViT follows the spirit of the image model ``Vision Transformer'' (ViT~\cite{dosovitskiy2020image}), but extends its self-attention mechanism from the image space to the space-time-modalitiy 4D volume. 
%
In particular, MM-ViT operates in the compressed video domain, and thus it could exploit the readily available visual modalities, i.e. \textit{I-frames}, \textit{motion vectors} and \textit{residuals}, to model appearance and temporal structure, and further leverage \textit{audio} content for additional supervision (see Figure.~\ref{Fig:preview}). 
In order to maintain a reasonable computational cost, we design several efficient model variants which factorize the self-attention computation along the space, time and modality dimensions. 
Further, three distinct cross-modal attention mechanisms are proposed to facilitate learning the inter-modal interactions. 
We extensively evaluate our model on three action recognition datasets, and demonstrate that MM-ViT outperforms alternative video transformer architectures, and performs comparably, and in some cases superior, to the established CNN counterparts with the assistance of computationally expensive optical flow. 
\section{Related Work}
\noindent \textbf{Video Action Recognition: }
The tremendous success of deep learning methods on image-based recognition tasks has inspired significant advance in video action recognition.
Most state-of-the-art solutions are based on CNNs and can be broadly classified into 2D- and 3D-CNN methods. 
The 2D-CNN approaches~\cite{feichtenhofer2016convolutional,jiang2019stm,karpathy2014large,li2020tea,simonyan2014two,lin2019tsm,ullah2017action,wang2016temporal,zhou2018temporal} process individual frames to extract frame-level features and then aggregate them with some kind of temporal modeling including late-fusion/mid-level fusion~\cite{karpathy2014large,wang2016temporal,zhou2018temporal}, LSTM~\cite{li2017skeleton,ullah2017action}, channel-shift~\cite{lin2019tsm}, feature temporal differencing~\cite{jiang2019stm,li2020tea}. 
Alternative multi-stream formulations add additional CNN streams to incorporate auxiliary data modalities, e.g., optical flow~\cite{fan2019more,feichtenhofer2016convolutional,kwon2020motionsqueeze,simonyan2014two},  audio~\cite{gao2020listen,kazakos2019epic}, human skeleton estimates~\cite{du2015hierarchical,yan2018spatial}. 
In contrast, 3D-CNN methods~\cite{carreira2017quo,feichtenhofer2020x3d,feichtenhofer2019slowfast,hara2018can,ji20123d,tran2015learning} jointly learn spatial-temporal features by conducting convolution over a sequence of frames. 
Although better accuracy can be achieved, 3D convolution are computationally heavy, making the
deployment difficult. 
To alleviate this problem, recent research~\cite{chen2020residual,qiu2017learning,tran2019video,tran2018closer} proposes a family of pseudo-3D convolutions to reduce computational cost while preserve accuracy. 

\noindent \textbf{Compressed Video Action Recognition:} 
Hand crafted motion representation, i.e. dense optical flow, contains complementary temporal information to the deep learning features, and thus employing optical flow almost guarantees to improve the recognition performance~\cite{fan2019more,kwon2020motionsqueeze,lin2019tsm,qiu2019learning,tran2018closer}. 
However, optical flow is unfortunately expensive to compute. 
To bypass this problem, pioneering works~\cite{zhang2016real,zhang2018real} replace optical flow with \textit{motion vectors} which present in the compressed video for encoding the movements of pixel blocks. 
Following up works achieve better performance by further exploiting \textit{residuals}~\cite{wu2018compressed} or refining motion vectors with the supervision of optical flow during training~\cite{shou2019dmc}.
Despite obtaining promising results, unlike our model, these methods ignore the rich inter-modal relations and fail to leverage \textit{audio} signal. 

\noindent \textbf{Deep Multi-Modal Representation Learning:}
Riding on the success of deep learning, many deep multi-modal representation learning approaches have been proposed. 
Some suggest to fuse features from different modalities in a joint latent space by bilinear pooling~\cite{fukui2016multimodal}, outer-product~\cite{zadeh2017tensor}, and statistical regularization~\cite{aytar2017cross}. 
Inspired by the strong progress made in learning contextualized language representations using transformers~\cite{devlin2018bert,liu2019roberta,sanh2019distilbert,yang2019xlnet}, several recent works have extended uni-modal transformers to the multi-modal setting by adding additional transformer stream(s) to learn joint vision-language representations~\cite{luo2021clip4clip,radford2021learning}, vision-audio representations~\cite{lee2020parameter} and vision-language-audio representations~\cite{tsai2019multimodal,zadeh2019factorized}.
Our model is also built on the transformer, while it differs to the previous works by using a single transformer to model both intra- and inter-modal features with the help of effective cross-modal attention. 

\noindent \textbf{Vision Transformer:}
Self-attention has been extensively employed in computer vision tasks, e.g., image classification~\cite{bello2019attention,dosovitskiy2020image,liu2021swin}, object detection~\cite{carion2020end,hu2018relation,zhu2020deformable}, and video classification~\cite{chen20182,girdhar2019video,wang2018non}.
Typically, it is used in conjunction with CNNs to augment convolutional features~\cite{bello2019attention,carion2020end,hu2018relation,neimark2021video,zhu2020deformable}. 
Until recently, Vision Transformer (ViT)~\cite{dosovitskiy2020image} proposes a pure transformer architecture replacing all convolutions with self-attention, and it outperforms CNN counterparts in several downstream image tasks. 
Built on the ViT design,  Timesformer~\cite{bertasius2021space} and ViViT~\cite{arnab2021vivit} extend 2D spatial self-attention to the 3D spatial-temporal volume for video classification. 
In order to maintain a manageable computational cost, both works propose model variants which factorize different components of the transformer over the spatial- and temporal-dimension. 
In this paper, we present a new pure-transformer approach for video action recognition.
Similar to~\cite{arnab2021vivit,bertasius2021space}, our architecture stems mainly from the ViT design.
However, instead of using solely decoded RGB frames, our model directly applies to the compressed videos and exploits both visual and audio modalities in the compressed domain.
We empirically demonstrate that MM-ViT outperforms Timesformer~\cite{bertasius2021space} and ViViT~\cite{arnab2021vivit} on three public action recognition benchmarks, while being more computationally efficient (in terms of FLOPs). 
\section{Methodology}
\subsection{Representations in Compressed Videos}
Common video compression algorithms, such as MPEG-4, H.264 and HEVC, enable highly efficient video compression by spliting a video into {\it I-frames} (intra-coded frames), {\it P-frames} (predictive frames) and/or {\it B-frames} (bi-directional frames).  
An {\it I-frame} is a regular image and compressed as such. A {\it P-frame} holds only changes in the image from the previous frame, thus saving space. 
%
In practice, a {\it P-frame} comprises \textit{motion vectors} and a \textit{residual}. 
The \textit{motion vectors} represent the movements of block of pixels (typically a frame is divided into 16x16 macroblocks during video compression) from the source frame to the target frame, thus roughly resemble coarse-grained optical flows. 
A \textit{residual} retains the RGB pixel difference
between a \textit{P-frame} and its reference \textit{I-frame} after
motion compensation based on motion vectors, therefore, it usually shows changes of appearance and motion boundaries.  
A \textit{B-frame} can be regarded as a special \textit{P-frame}, where \textit{motion vectors} are computed bi-directionally. 
In addition to the visual modalities, a compressed video also includes an audio stream which, in many cases, could play a critical role in recognition~\cite{gao2020listen,nagrani2020speech2action}. 
%


\subsection{Embedding Compressed Video Clips}
MM-ViT operates on a compressed video clip $\mathcal{V}$. The vision modalities consist of $T$ sampled {\it I-frames}, {\it motion vectors} and {\it residuals} of height $H$ and width $W$.
The audio modality contains a 1D waveform of length $T'$. 
Formally, $\mathcal{V} = \{ \mathcal{I} \in \mathbb{R}^{T \times 3 \times H \times W },   \mathcal{M} \in \mathbb{R}^{T \times 2 \times H \times W }, \mathcal{R} \in \mathbb{R}^{T \times 3 \times H \times W }, \mathcal{A} \in \mathbb{R}^{T'}  \}$,
where $\mathcal{I}$, $\mathcal{M}$, $\mathcal{R}$ and $\mathcal{A}$ represents  {\it I-frame}, {\it motion vector}, {\it residual} and {\it audio} modality, respectively. 
In order to (roughly) align the visual and audio signals, we partition the 1D audio waveform to $T$ segments and project each segment to a 128-dimensional vector using a pretrained VGGish model~\cite{gemmeke2017audio}.
%
 
%

Following ViT~\cite{dosovitskiy2020image}, we decompose each RGB {\it I-frame} into $N$ non-overlapping patches of size $P \times P$. Then, we project those patches into token embeddings using a learnable linear embedding layer $\mathbf{E}^{\mathcal{I}} \in \mathbb{R}^{d \times 3P^2}$.
Additionally, a spatiotemporal positional encoding $PE^{\mathcal{I}}_{(p, t)} \in \mathbb{R}^d$ is added to each patch token in order to preserve the positional information. 
Same operations are applied to tokenize the \textit{motion vectors} and \textit{residuals} as well:
\begin{align}
z^{\mathcal{I}}_{(p, t)} &= \mathbf{E}^{\mathcal{I}} \cdot  \mathcal{I}_{(p, t)} + PE^{\mathcal{I}}_{(p, t)} \\
z^{\mathcal{M}}_{(p, t)} &= \mathbf{E}^{\mathcal{M}} \cdot  \mathcal{M}_{(p, t)} + PE^{\mathcal{M}}_{(p, t)} \\
z^{\mathcal{R}}_{(p, t)} &= \mathbf{E}^{\mathcal{R}} \cdot  \mathcal{R}_{(p, t)} + PE^{\mathcal{R}}_{(p, t)} 
\end{align}
where $z^{\mathcal{I}}_{(p, t)}, z^{\mathcal{M}}_{(p, t)}, z^{\mathcal{R}}_{(p, t)}$ are the resulting vision tokens ($p = 1,...,N, \quad t=1,..,T$). 
For the audio feature, we first apply a linear layer $\mathbf{E}^{\mathcal{A}} \in \mathbb{R}^{d \times 128} $ to project it to the same dimensional space as vision tokens,  then a temporal positional encoding $PE^{\mathcal{A}}_{(t)} $ is added:
\begin{align}
z^{\mathcal{A}}_{(t)} &= \mathbf{E}^{\mathcal{A}} \cdot \Phi(\mathcal{A}_{(t)}) + PE^{\mathcal{A}}_{(t)} 
\end{align}
where the transformation function $\Phi$ is parameterized by the VGGish model~\cite{gemmeke2017audio}. 
To facility fully spatiotemporal self-attention across visual and audio modalities, we replicate each audio token $z^{\mathcal{A}}_{(t)} $ $N$ times along the spatial dimension, thus  $z^{\mathcal{A}}_{(p,t)} = z^{\mathcal{A}}_{(p',t)}$ for $p, p' \in \{1,...,N \}$.

The resulting token sequences $z^{\mathcal{I}}_{(p, t)}, z^{\mathcal{M}}_{(p, t)}, z^{\mathcal{R}}_{(p, t)}$ and $z^{\mathcal{A}}_{(p, t)}$ for $p = 1,...,N, t=1,...,T$, and a special ``{\normalfont CLS}'' token  $z_{(0,0)}^{{\normalfont CLS}}$ constitute the input to MM-ViT. 
Following BERT~\cite{devlin2018bert}, we use the output embedding of $z_{(0,0)}^{{\normalfont CLS}}$ as the aggregated representation for the entire input sequence. 


\begin{figure*}
\begin{center}
\includegraphics[width=1.0\linewidth]{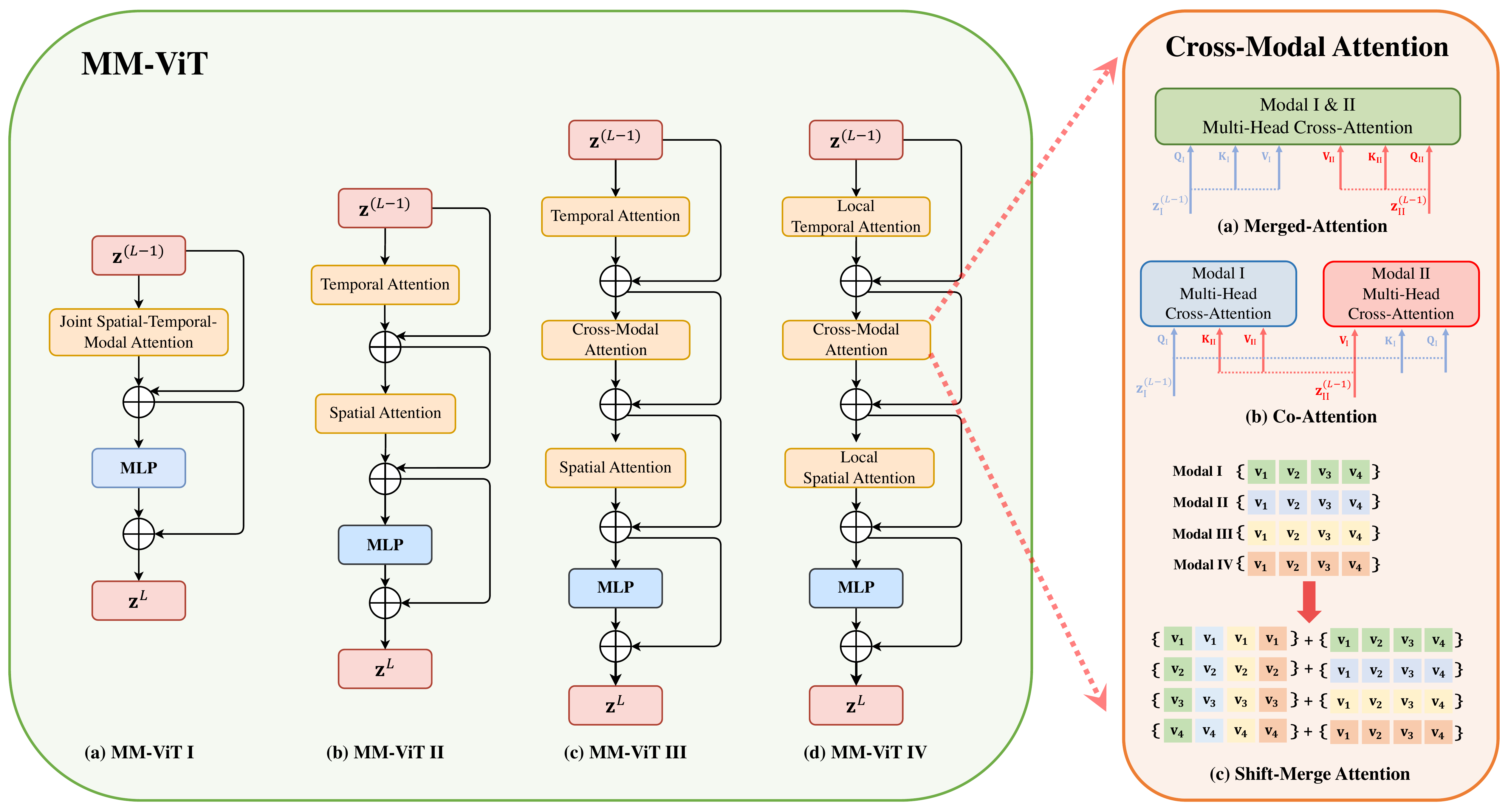}
\end{center}
\caption{
Overview of the investigated self-attention blocks in MM-ViT and the proposed cross-modal attention mechanisms. 
%
%
}
%
%
\label{Fig:arch}
\vglue -0.4cm
\end{figure*}

\subsection{Multi-modal Video Transformers}
%

In this section, we introduce four multi-modal video transformer architectures.
We begin with an architecture which simply adopts the standard self-attention mechanism to measure all pairwise token relations. 
We then present three efficient model variants, which factorize the self-attention computation over the space-time-modality 4D volume with distinct strategies (see Figure.~\ref{Fig:arch}).

\noindent \textbf{MM-ViT I - Joint Space-Time-Modality Attention}: 
Similar in spirit to the ``Joint Space-Time Attention'' in Timesformer~\cite{bertasius2021space} and the ``Spatio-Temporal Attention'' in ViViT~\cite{arnab2021vivit}, each transformer layer of this model measures pairwise interactions between all input tokens. 
Concretely, MM-ViT I consists of $L$ transformer layers. At each layer, a set of query($q$), key($k$) and value($v$) vectors are first computed for every input token embedding $z^{(l-1, s)}_{(p, t)}$ from the preceding layer as below:
\begin{align}
q^{(l, s)}_{(p, t)} &= \mathbf{W}^{l}_{Q} \cdot \mathrm{LN}(z^{(l-1, s)}_{(p, t)}) \\
k^{(l, s)}_{(p, t)} &= \mathbf{W}^{l}_{K} \cdot \mathrm{LN}(z^{(l-1, s)}_{(p, t)}) \\
v^{(l, s)}_{(p, t)} &= \mathbf{W}^{l}_{V} \cdot \mathrm{LN}(z^{(l-1, s)}_{(p, t)})
\end{align}
where $\mathbf{W}^{l}_{Q}, \mathbf{W}^{l}_{K}, \mathbf{W}^{l}_{V} \in \mathbb{R}^{d_h \times d}$ are learnable embedding matrices, {\normalfont LN($\cdot$)} denotes layer normalization~\cite{ba2016layer}, $s \in \mathbb{S}=\{ \mathcal{I}, \mathcal{M}, \mathcal{R}, \mathcal{A} \}$.
Then, the self-attention weights for query patch $q^{(l, s)}_{(p, t)}$ are given by:
\begin{align}
\alpha^{(l, s)}_{(p, t)} = \mathrm{Softmax}\Bigl(\frac{q^{(l, s)}_{(p, t)}}{\sqrt{d_{h}}} \cdot \Bigl\{k^{(l,s^{'})}_{(p^{'},t^{'})} \Bigr\}_{\substack{p^{'} = 0,...,N, s^{'} = \mathbb{S} \\ t^{'} = 0,...,T  }} \Bigr)
\end{align}
The output token  $z^{(l, s)}_{(p, t)}$ is further obtained by first computing the weighted sum of the value vectors based on the self-attention weights, followed by a linear projection through a MLP block. \textit{Residual connection} is employed to promote robustness. 

In practice, we adopt the Multi-Head Self-Attention (MSA) which yields better performance. Specifically, MSA uses $h$ sets of $\{\mathbf{W}^{l}_{Q}, \mathbf{W}^{l}_{K}, \mathbf{W}^{l}_{V} \}$.
The outputs of the $h$ heads are concatenated and forwarded to the next layer.
%
Note that although this model allows interactions between all token pairs, it has quadratic computational complexity with respect to the number of  tokens. 

\noindent \textbf{MM-ViT II - Factorized Space-Time Attention}:
Instead of computing self-attention across all pairs of input tokens, this model factorizes the operation along the spatial- and temporal-dimensions. 
As shown in Figure~\ref{Fig:arch} (``MM-ViT II'' in the left panel), given an token $z^{(l-1, s)}_{(p, t)}$ from layer $l-1$, we first conduct self-attention temporally (or spatially) by comparing it with all tokens at the same spatial location across all modalities.  
Next, a spatial attention followed by a linear projection is applied to generate the output embedding $z^{(l,s)}_{(p, t)}$ from layer $l$.
Formally, we define our factorized space-time attention as:
\begin{align}
y^{(l, s)}_{(p, t)} &= \mathrm{MSA}_{\mathrm{time}}\bigl(\mathrm{LN}(z^{(l-1, s)}_{(p, t)})\bigr) + z^{(l-1, s)}_{(p, t)} \\
y^{' (l, s)}_{(p, t)} &= \mathrm{MSA}_{\mathrm{space}}\bigl(\mathrm{LN}(y^{(l, s)}_{(p, t)})\bigr) + y^{(l, s)}_{(p, t)} \\
z^{(l,s)}_{(p, t)} &= \mathrm{MLP}\bigl(\mathrm{LN}(y^{' (l, s)}_{(p, t)})\bigr) + y^{' (l, s)}_{(p, t)} 
\end{align}
%
%
This architecture introduces more parameters than MM-ViT I due to one additional MSA operation. 
However, by decoupling self-attention over the input spatial- and temporal-dimensions,  MM-ViT II reduces computational complexity per patch from $\mathcal{O}(N \cdot T \cdot |\mathbb{S}|)$ to $\mathcal{O}(N \cdot |\mathbb{S}| + T \cdot |\mathbb{S}|)$.

\noindent \textbf{MM-ViT III - Factorized Space-Time Cross-Modal Attention:}
Our third model further factorizes self-attention over the modality dimension. 
At each transformer layer, it attends to space, time and modality dimensions sequentially, thus reducing the computational complexity per patch to  $\mathcal{O}(N + T + |\mathbb{S}|)$.  
Concretely, a patch token $z^{(l,s)}_{(p, t)}$ from layer $l$ is calculated as follows:
\begin{align}
y^{(l, s)}_{(p, t)} &= 
\mathrm{MSA}_{\mathrm{time}} \bigl(\mathrm{LN}(z^{(l-1, s)}_{(p, t)})\bigr) + z^{(l-1, s)}_{(p, t)}\\
y^{' (l, s)}_{(p, t)} &= 
\mathrm{MCA} \bigl(\mathrm{LN}(y^{(l, s)}_{(p, t)})\bigr) + y^{(l, s)}_{(p, t)}\\
y^{'' (l, s)}_{(p, t)} &= \mathrm{MSA}_{\mathrm{space}}\bigl(\mathrm{LN}(y^{'(l, s)}_{(p, t)})\bigr) + y^{'(l, s)}_{(p, t)}\\
z^{(l,s)}_{(p, t)} &= \mathrm{MLP}\bigl(\mathrm{LN}(y^{'' (l, s)}_{(p, t)})\bigr) + y^{'' (l, s)}_{(p, t)} 
\end{align}
where $\mathrm{MCA}$ stands for Multi-Head Cross-Attention, which is specifically designed for modeling cross-modal relations. 

The main idea for this architecture is to have an effective cross-modal attention to facilitate learning from multi-modal data. 
To this end,  we develop three different cross-model attention mechanisms (see the right panel in Figure.~\ref{Fig:arch}).
%
The first one is called ``Merged-Attention". 
Given a query from one modality, it considers all of the keys and values regardless of the modality type.  
The output of this cross-attention module for query $q^{(l,s)}_{(p, t)}$ is defined as:
\begin{multline}
\mathrm{A}(q^{(l,s)}_{(p, t)}, \mathbf{K}_{(p,t)}^{(l,\mathbf{s}')}, \mathbf{V}_{(p,t)}^{(l, \mathbf{s}')})= \\
\mathrm{Softmax}\bigl( \frac{q^{(l,s)}_{(p, t)} \cdot \mathbf{K^\top}_{(p,t)}^{(l,\mathbf{s}')}}{\sqrt{d_h}} \bigr) \cdot  \mathbf{V}_{(p,t)}^{(l, \mathbf{s}')}
\end{multline}
where $\mathbf{s}' = \mathbb{S}$.
Alternatively, one can allow queries to interact only with keys and values from other modalities, thus $\mathbf{s}' = \mathbb{S} \setminus \{s\}$.
We refer this cross-modal attention to ``Co-Attention". 
Note that it is identical to the cross-attention operation proposed for the decoder model in Transformer~\cite{vaswani2017attention}.
%


Lastly, we propose a computation-free shift-based method called ``Shift-Merge Attention'' to assist interaction across modalities.  
It shares a similar spirit to the shift approaches proposed in the CNN domain (e.g., TSM~\cite{lin2019tsm}), seeking to strike a balance between accuracy and efficiency. 
More specifically, we discard queries and keys, and directly work on the value embeddings by first evenly splitting each $v^{(l,s)}_{(p, t)}$ into four portions ${v_1}^{(l,s)}_{(p, t)}, {v_2}^{(l,s)}_{(p, t)}, {v_3}^{(l,s)}_{(p, t)}$, and ${v_4}^{(l,s)}_{(p, t)}$. 
Then, we shift and mix the value embedding portions from different modalities, but at the same spatial and temporal index as follows:
\begin{small}
\begin{align}
{r}^{(l,\mathcal{I})}_{(p, t)} &= {v_{1}}^{(l,\mathcal{I})}_{(p, t)} \parallel
{v_{1}}^{(l,\mathcal{M})}_{(p, t)} \parallel
{v_{1}}^{(l,\mathcal{R})}_{(p, t)} \parallel
{v_{1}}^{(l,\mathcal{A})}_{(p, t)}  + {v}^{(l,\mathcal{I})}_{(p, t)}
\\
{r}^{(l,\mathcal{M})}_{(p, t)} &= {v_{2}}^{(l,\mathcal{I})}_{(p, t)} \parallel
{v_{2}}^{(l,\mathcal{M})}_{(p, t)} \parallel
{v_{2}}^{(l,\mathcal{R})}_{(p, t)} \parallel
{v_{2}}^{(l,\mathcal{A})}_{(p, t)} +
{v}^{(l,\mathcal{M})}_{(p, t)}
\\
{r}^{(l,\mathcal{R})}_{(p, t)} &= {v_{3}}^{(l,\mathcal{I})}_{(p, t)} \parallel
{v_{3}}^{(l,\mathcal{M})}_{(p, t)} \parallel
{v_{3}}^{(l,\mathcal{R})}_{(p, t)} \parallel
{v_{3}}^{(l,\mathcal{A})}_{(p, t)} +
{v}^{(l,\mathcal{R})}_{(p, t)}
\\
{r}^{(l, \mathcal{A})}_{(p, t)} &={v_{4}}^{(l,\mathcal{I})}_{(p, t)} \parallel
{v_{4}}^{(l,\mathcal{M})}_{(p, t)} \parallel
{v_{4}}^{(l,\mathcal{R})}_{(p, t)} \parallel
{v_{4}}^{(l,\mathcal{A})}_{(p, t)} +
{v}^{(l,\mathcal{A})}_{(p, t)}
\end{align}
\end{small}
where ${r}$ denotes the resulting encoding, $\parallel$ represents concatenation. 
We also add a \textit{residual connection} to preserve the learning capability.

\noindent \textbf{MM-ViT IV - Factorized Local Space-Time Cross-Modal Attention}:
The last proposed architecture restricts the factorized spatial and temporal attention in MM-ViT III to non-overlapping local windows, therefore,  the computational cost is further reduced. 
Supposing a local spatial and temporal window contains $M$ and $F$ patches, respectively,  the computational complexity per patch becomes $\mathcal{O}(M + F + |\mathbb{S}|)$. In our experiments, we set $M = \frac{N}{4}, F = \frac{T}{2}$.  
However, limiting the receptive field to a local window may adversely affect model's performance. 
To alleviate this issue, we propose to insert a convolution layer after the temporal and spatial attention to strengthen the connection between neighboring windows. 
The convolution kernel size is same as the window size, and the stride size = 1. 

Please note we are aware of that the order of the spatial, temporal and cross-modal attention may have an impact to model performance, and leave a further discussion to Sec.~\ref{sec:AS}.

\section{Experimental Evaluation}
In this section, we first describe our experimental setup,
then present ablation studies about some design choices, and finally compare with state-of-the-art methods.
%
%
%
\subsection{Setup}
\noindent \textbf{Datasets:}  
We evaluate the proposed method on three popular video action recognition datasets: UCF101~\cite{soomro2012ucf101}, Something-Something-V2~\cite{goyal2017something} and Kinetics-600~\cite{kay2017kinetics}.
UCF-101 contains 13,320 trimmed short videos from 101 action categories. It has three training-testing splits. We report the average performance across the three splits. 
Kinetics-600 contains around 480K 10-seconds long videos for 600 action classes. 
Something-Something-v2 (SSv2) consists of about 220K videos with a time span from 2 to 6 seconds for 174 action classes. 
Different to other datasets, it places more emphasize on a model's ability to recognize fine-grained actions since the same background scenes can be shared aross many classes. 
In addition, the released version of SSv2 has no audio stream, therefore,  we only extract the visual modalities from SSv2 to evaluate the proposed models.
%
%
In our experiments, we convert all compressed videos to MPEG-4 codec which encodes a video into \textit{I-frames} and \textit{P-frames}. In average, an \textit{I-frame} is followed by 11 \textit{P-frames}.

\noindent \textbf{Training Details:}
Following~\cite{shou2019dmc,wu2018compressed}, we resize all training videos to  $340\times 256$.
Then, random horizontal flipping (omitted for SSv2) and random cropping ($224\times 224$) are applied to \textit{I-frames}, \textit{motion vectors} and \textit{residuals} for data augmentation.
Patch size is set to $16\times 16$ across the visual modalities.
The audio waveform is partitioned to 1-second long segments and projected into 128-dimensional vectors by VGGish~\cite{gemmeke2017audio}.
We use ViT-B/16~\cite{dosovitskiy2020image} pretrained on ImageNet-21K as our backbone, and fine-tuned using SGD with a batch size of 8. 
Learning rate starts from 0.5 and is divided by 10 when the validation accuracy plateaus. 

\noindent \textbf{Inference Details:}
During inference, unless otherwise mentioned, the input consists of 8 uniformly sampled triplets of \textit{I-frames}, \textit{motion vectors}, \textit{residuals} with a crop size of $224\times 224$,  and audio features (omitted for SSv2) that are temporally aligned with the visual features. 
We closely follow~\cite{bertasius2021space} to report accuracy from three spatial crops (left, centre, right) and average the scores for final prediction. 
%
%

%
%
%

\subsection{Ablation Studies} 
\label{sec:AS}
\noindent \textbf{Analysis of the proposed model variants:}
In this section, we first compare the performance of the proposed model variants on UCF-101 and SSv2, in terms of accuracy and efficiency. 
Table~\ref{tbl:UCF101_comp} summarizes the detailed experimental results.
%
%
%
We first note that MM-ViT I underperforms to the factorized alternatives (MM-ViT II $\&$ III), although it consumes more computational cost.
This is consistent with the observation in~\cite{bertasius2021space},
and we suspect it is due to the lack of dedicated parameters to model spatial, temporal and cross-modal attentions separately.
Moreover, our results show that factorizing self-attention over the input dimensions consistently improve both recognition accuracy and  efficiency, e.g., MM-ViT II outperforms MM-ViT I by 0.83$\%$ on UCF-101 while incurs 32$\%$ less FLOPs, 
meanwhile, MM-ViT III (Merged-Att.) outperforms MM-ViT II by 1.75$\%$ in accuracy and requires 3.5$\%$ less FLOPs.

Among the three proposed cross-modal attention mechanisms, ``Merged-Attention'' achieves the best accuracy on both UCF-101 and SSv2. 
This suggests that sharing keys and values across all modalities is critical to obtain a comprehensive understanding of the video content. 
Interestingly, the ``Shift-Merge Attention'' performs comparably to the ``Merged-Attention'' while being more efficient, making it attractive in resource constrained scenarios. 
When restricting the self-attention to local views (MM-ViT IV), the accuracy has a significant drop ($\downarrow$3.11$\%$ on UCF101, $\downarrow$4.72$\%$ on SSv2),
indicating more sophisticated cross-window connection may be needed to mitigate the information loss from using local attention views.

%

\begin{table}[!thb]
\caption{Performance comparison of the proposed model variants on UCF101 and SSv2. Note we report FLOP numbers for UCF-101 where both visual and audio modalities are involved.}
\vglue -0.30cm
\label{tbl:UCF101_comp}
\centering
\setlength\extrarowheight{1pt}
\resizebox{1\columnwidth}{!}{%
\begin{tabular}{ c c c c c}
\Xhline{1pt}
\multirow{1}{*}{\textbf{Model}}	
&   \multicolumn{1}{c}{\textbf{Params}} 
&   \multicolumn{1}{c}{\textbf{TFLOPs}} 
&   \multicolumn{1}{c}{\textbf{UCF-101}} 
&   \multicolumn{1}{c}{\textbf{SSv2}} 
\\
\hline		
MM-ViT I & 87.23M & 1.26 & 93.41 & 62.55\\
\hline
MM-ViT II & 122.69M & 0.85 & 94.24 & 62.05\\
\hline
MM-ViT III (Merged-Att.) & 158.14M & 0.82 & \textbf{95.99} & \textbf{64.97}  \\
\hline
MM-VIT III (Co-Att.) & 158.14M & 0.82 & 93.87 & 62.10 \\
\hline
MM-VIT III (Shift-Merge Att.) & 143.98M & 0.78 & 95.04 & 63.55 \\
\hline
MM-ViT IV (Merged-Att.) & 158.60M & 0.72 & 92.88 & 60.25 
\\
\Xhline{1pt}
\end{tabular}}
\end{table}

\noindent \textbf{Effect of attention order:}
%
%
In this part, we evaluate the effect of attention order by enumerating all possible orders of the spatial, temporal and cross-modal attention. 
For simplicity, we only report the results from MM-ViT III with Merged-Attention on UCF-101.
%
%
As shown in Table~\ref{tbl:attn_order}, conducting temporal attention before spatial attention slightly, but consistently, performs better than the opposite, 
which can be explained as temporal attention provides key clues for distinguishing actions that share similar appearance features.
%
%
In particular, the best performing attention order is \textit{temporal}$\Rightarrow$ \textit{cross-modal}$\Rightarrow$\textit{spatial}. 
%


\begin{table}[!thb]
\vglue -0.20cm
\caption{Effect of attention order on UCF101. T, S and M represent temporal, spatial and cross-modal attention, respectively.}
\vglue -0.30cm
\label{tbl:attn_order}
\centering
\setlength\extrarowheight{1pt}
\resizebox{0.7\columnwidth}{!}{%
\begin{tabular}{ c c c}
\Xhline{1pt}
\multirow{1}{*}{\textbf{Attention Order}}	
&   \multicolumn{1}{c}{\textbf{Top-1}} 
&   \multicolumn{1}{c}{\textbf{Top-5}} 
\\
\hline		
T $\Rightarrow$ S $\Rightarrow$ M & 95.41 & 99.39  \\
\hline
S $\Rightarrow$ T $\Rightarrow$ M & 95.24 & 99.40 \\
\hline
T $\Rightarrow$ M $\Rightarrow$ S  & \textbf{95.99} & \textbf{99.54} \\
\hline
S $\Rightarrow$ M $\Rightarrow$ T  & 95.31 & 99.50 \\
\hline
M $\Rightarrow$ T $\Rightarrow$ S  & 95.00 & 99.21 \\
\hline
M $\Rightarrow$ S $\Rightarrow$ T  & 94.85 & 99.25 \\
\Xhline{1pt}
\end{tabular}}
\vglue -0.2cm
\end{table}

\noindent \textbf{Effect of input modality:}
%
%
To evaluate the importance of each data modality, we conduct an ablation study by training and evaluating the best performing model (i.e. MM-ViT III with ``Merged-Attention'') with different modality combinations on UCF-101.
As shown in Table~\ref{tbl:modality_ablation}, \textit{I-frame} is the most essential data modality as removing it alone decreases top-1 accuracy by 4.11$\%$. 
The \textit{motion vector} and \textit{residual frame} also play important roles for video recognition.
Without either modality can lead to an accuracy drop up to 2.54$\%$.
%
%
It is interesting that the audio modality has a major impact to video recognition as well, which is confirmed by the significant performance degradation (2.42$\%$ drop in top-1 accuracy) when excluding audio input. 
It is likely because audio contains dynamics and contextual temporal information that is beneficial for action recognition (e.g., the sound of axe hitting tree is discriminative for recognizing ``Cutting Tree'' ). 

\begin{table}[!thb]
\vglue -0.20cm
\caption{Performance comparison of different modality combinations on UCF-101.}
\vglue -0.30cm
\label{tbl:modality_ablation}
\centering
\setlength\extrarowheight{1pt}
\resizebox{0.7\columnwidth}{!}{%
\begin{tabular}{ c c c c | c c }
\Xhline{1pt}
\multirow{1}{*}{\textbf{$\mathcal{I}$}}	
&   \multicolumn{1}{c}{\textbf{$\mathcal{M}$}} 
&   \multicolumn{1}{c}{\textbf{$\mathcal{R}$}} 
&   \multicolumn{1}{c}{\textbf{$\mathcal{A}$}}
&   \multicolumn{1}{c}{\textbf{Top-1}} 
&   \multicolumn{1}{c}{\textbf{Top-5}} 
\\
\hline
\cmark & \cmark & \cmark & \cmark & 95.99$\%$ & 99.54$\%$ \\
\xmark & \cmark & \cmark & \cmark & 91.88$\%$ & 98.76$\%$ \\
\cmark & \xmark & \cmark & \cmark & 93.45$\%$ & 99.20$\%$ \\
\cmark & \cmark & \xmark & \cmark & 94.16$\%$ & 99.28$\%$ \\ 
\cmark & \cmark & \cmark & \xmark & 93.57$\%$ & 99.21$\%$ \\
\Xhline{1pt}
\end{tabular}}
\vglue -0.2cm
\end{table}



\subsection{Comparison to State-of-the-Art}
In this section, we compare our best performing architecture, i.e. MM-VIT III with ``Merged Attention'', to other state-of-the-art approaches on UCF101, SSv2 and Kinetics-600 datasets.
Unless otherwise specified, we follow~\cite{bertasius2021space} to report results from $1\times 3$ views (1 temporal and 3 spatial crops).

\noindent \textbf{UCF101:}
Table~\ref{tbl:ucf101} summarizes the performance of our model and other competing methods on UCF-101. 
%
We first observe that even without using audio, MM-ViT already outperforms all other methods that operate the same compressed video modalities~\cite{zhang2016real,zhang2018real,wu2018compressed,shou2019dmc}, by up to $6.9\%$ in top-1 accuracy.
It suggests that the explicit reasoning of inter-modal relations in MM-ViT is effective. 
Further improvements are achieved by incorporating audio signal ($\uparrow$ 2.1$\%$ in top-1 accuracy) and pre-training the model on Kinetics-600 ($\uparrow$ 3.5$\%$ in top-1 accuracy). 
Remarkably, we surpass all CNN alternatives with or without optical flow, and thus establishes a new state-of-the-art for UCF-101.

%
%

%

\begin{table}[!thb]
\vglue -0.20cm
\caption{Performance comparison with the state-of-the-art methods on UCF-101.}
\vglue -0.30cm
\label{tbl:ucf101}
\centering
\setlength\extrarowheight{1pt}
\resizebox{1.0\columnwidth}{!}{%
\begin{tabular}{ c c c c}
\Xhline{1pt}
\multirow{1}{*}{}	
&   \multicolumn{1}{c}{\textbf{Modality}}
&   \multicolumn{1}{c}{\textbf{Pretrain}} 
&   \multicolumn{1}{c}{\textbf{Top-1}} 
\\
\hline		
C3D~\cite{tran2015learning} & RGB & ImageNet1K & 82.3 \\
ActionFlowNet~\cite{ng2018actionflownet} & RGB & - & 83.9 \\
Res3D~\cite{tran2017convnet} & RGB & Sports-1M & 85.8 \\
PWC-Net(ResNet-18)~\cite{sun2018pwc} & RGB & - & 90.6 \\
ARTNet~\cite{wang2018appearance} & RGB & Kinetics &94.3 \\
I3D RGB~\cite{carreira2017quo} & RGB & Kinetics & 95.6 \\
MF-Net~\cite{chen2018multi} & RGB & Kinetics & 96.0 \\
S3D~\cite{xie2017rethinking} & RGB & Kinetics & 96.5 \\
\hline	
Two-stream~\cite{simonyan2014two} & RGB + Flow & - & 88.0 \\
Two-stream fusion~\cite{feichtenhofer2016convolutional} & RGB + Flow & ImageNet1K & 92.5 \\
TV-Net~\cite{fan2018end} & RGB + Flow & - & 94.5 \\
R(2+1)D~\cite{tran2018closer} & RGB + Flow & Kinetics & 97.3 \\
I3D~\cite{carreira2017quo} & RGB + Flow & Kinetics &  98.0 \\
\hline
EMV-CNN~\cite{zhang2016real} & $\mathcal{I}$ + $\mathcal{M}$ +$\mathcal{R}$ & ImageNet1K & 86.4  \\
DTMV-CNN~\cite{zhang2018real} &$\mathcal{I}$ + $\mathcal{M}$ +$\mathcal{R}$  & - & 87.5 \\
CoViAR~\cite{wu2018compressed} & $\mathcal{I}$ + $\mathcal{M}$ +$\mathcal{R}$ & ImageNet1K & 90.4 \\
DMC-Net(I3D)~\cite{shou2019dmc} & $\mathcal{I}$ + $\mathcal{M}$ +$\mathcal{R}$ & ImageNet1K & 92.3 \\
MM-ViT  & $\mathcal{I}$ + $\mathcal{M}$ +$\mathcal{R}$ & ImageNet1K & 93.3 \\
MM-ViT & $\mathcal{I}$ + $\mathcal{M}$ +$\mathcal{R}$ + $\mathcal{A}$  & ImageNet1K & 95.4 \\
\textbf{MM-ViT} & $\mathcal{I}$ + $\mathcal{M}$ +$\mathcal{R}$ + $\mathcal{A}$ & Kinetics & \textbf{98.9} \\
\Xhline{1pt}
\end{tabular}}
\vglue -0.2cm
\end{table}

\noindent \textbf{SSv2:} 
In Table~\ref{tbl:SSv2}, we present detailed results on SSv2 including top-1$\&$top-5 accuracy, inference resolution and computational cost (in FLOPs). 
Notably, MM-ViT surpasses the Timesformer~\cite{bertasius2021space} and ViViT~\cite{arnab2021vivit} which also propose pure-transformer models, yet being more efficient in terms of inference FLOPs.
%
%
It confirms that the additional \textit{motion vector} and \textit{residual} modalities used by MM-ViT provide important complementary motion features, which could benefit the classification on ``temporally-heavy'' dataset like SSv2.
Furthermore, MM-ViT consistently performs better than the CNN counterparts that operate in the single RGB modality (by $\geq$ 1.5$\%$ in top-1 accuracy).
%
Although MM-ViT slightly underperforms to CNN-based MSNet-R50~\cite{kwon2020motionsqueeze} and blVNet~\cite{fan2019more} which use optical flow as an auxiliary modality, it eliminates the huge burden of optical flow computation and storage.

\begin{table}[!thb]
\caption{Performance comparison with state-of-the-art methods on SSv2. The inference resolution is denoted by $M \times T\times S^2$ for the number of modalities, temporal and spatial sizes.}
\vglue -0.30cm
\label{tbl:SSv2}
\centering
\setlength\extrarowheight{1pt}
\resizebox{1\columnwidth}{!}{%
\begin{tabular}{c c c c c c}
\Xhline{1pt}
\multirow{1}{*}{}	
&   \multicolumn{1}{c}{\textbf{Inference}} 
&   \multicolumn{1}{c}{\textbf{Top-1}} 
&   \multicolumn{1}{c}{\textbf{Top-5}} 
&   \multicolumn{1}{c}{\textbf{Inference}} 
\\
\multirow{1}{*}{} 
& \multicolumn{1}{c}{\textbf{Resolution}} 
& 
&
& \multicolumn{1}{c}{\textbf{TFLOPs$\times$views}} 
\\
\hline		
\textbf{CNN +  RGB only} & &\\
TRN~\cite{zhou2018temporal} & $ 1\times 8 \times 224^2$  &  48.8 & 77.6 & -\\
SlowFast~\cite{wu2020multigrid} & 
$1\times 64 \times 224^2$ & 61.7 & 87.8 & 0.07$\times$30\\
TSM~\cite{lin2019tsm} & $1\times 24 \times 224^2$ & 63.3 & 88.5 & 0.1$\times$1\\
STM~\cite{jiang2019stm} & 
$1\times 16 \times 224^2$ & 64.2 & 89.8 & 0.07$\times$10 \\
MSNet-R50~\cite{kwon2020motionsqueeze} & 
$1\times 16 \times 224^2$ & 64.7 & 89.4 & 0.07$\times$10 \\
TEA~\cite{li2020tea} & 
$1\times 16 \times 224^2$ & 65.1 & - & 0.07$\times$30 \\
blVNet~\cite{fan2019more} & 
$1\times 32 \times 224^2$ & 65.2 & 90.3 & 0.13$\times$30\\
\hline	
\textbf{CNN + RGB + Optical Flow} & &\\
TSM~\cite{lin2019tsm} &
$2\times 24 \times 224^2$ & 66.0 & 90.5 & - \\
MSNet-R50~\cite{kwon2020motionsqueeze} & 
$2\times 16 \times 224^2$ &  67.1 & 91.0 & -\\
\textbf{blVNet~\cite{fan2019more}} & 
$2\times 32 \times 224^2$  & \textbf{68.5} & \textbf{91.4} & -\\
\hline
\textbf{ Video Transformer} & & \\
Timesformer-L~\cite{bertasius2021space} & 
$1\times 96 \times 224^2$ & 62.4 & - & 2.38$\times$3\\
ViViT-L~\cite{arnab2021vivit} & 
$1\times 32 \times 224^2$ & 65.4 & 89.8 & 1.45$\times$4\\
MM-ViT & 
$3\times 8 \times 224^2$& 64.9 & 89.7 & 0.75$\times$3 \\
MM-ViT & 
$3\times 16 \times 224^2$ & 66.7& 90.8 & 1.51$\times$3\\
\Xhline{1pt}
\end{tabular}}
\vglue -0.2cm
\end{table}

\begin{figure*}[!thb]
	\centering
	\includegraphics[width=1.0\linewidth]{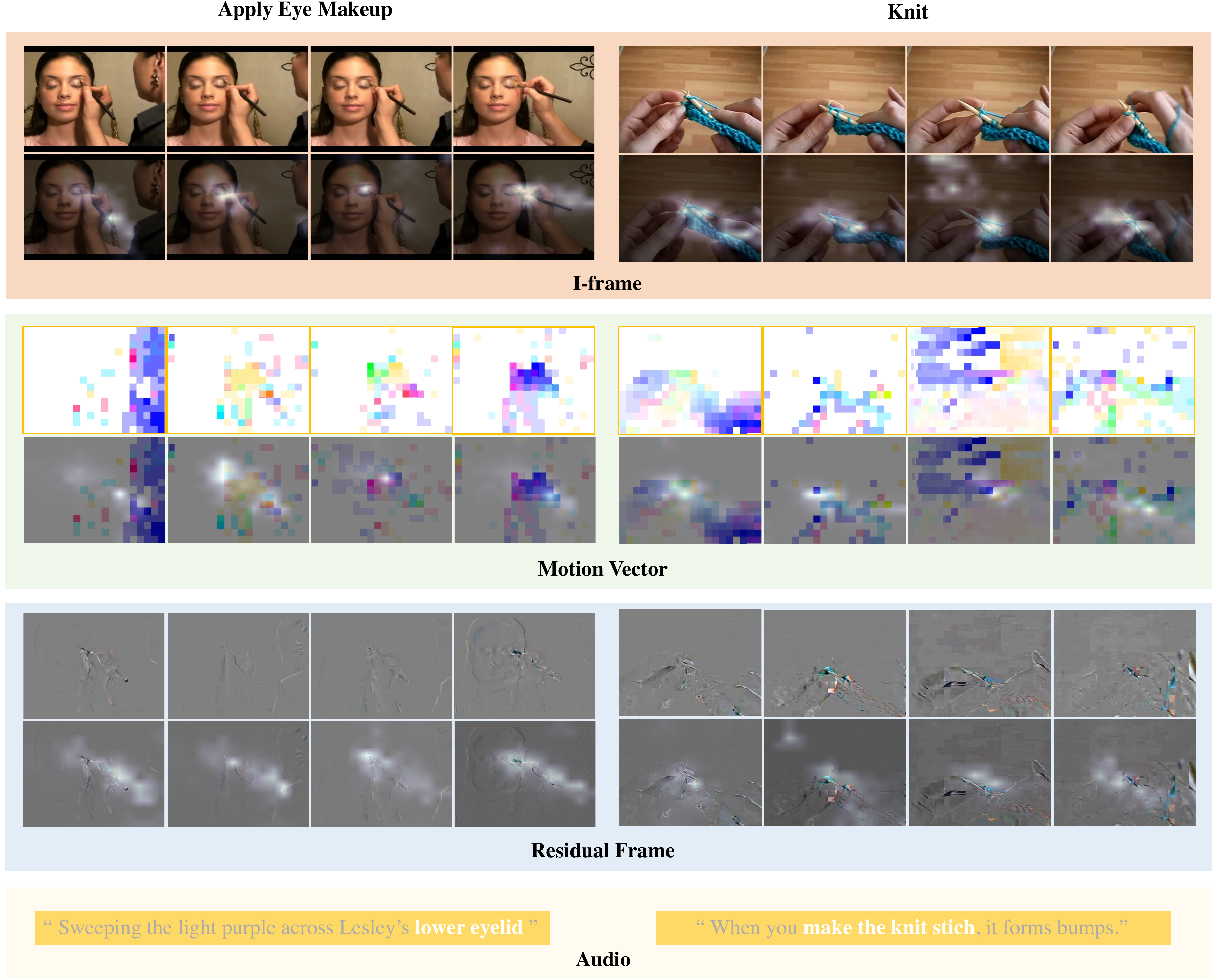}
	\caption{Visualization of attention weights to the input space. The proposed model learns to focus on relevant parts of the video for classification. }
	\label{fig:attn_vis}
	\vglue -0.3cm
\end{figure*}

\noindent \textbf{Kinetics-600:}
Kinetics-600 is a larger video classification dataset that shall further demonstrate the generalization of our approach. 
As shown in Table~\ref{tbl:kinetics600}, our MM-ViT  ($T=16$) achieves $83.5\%$ top-1 accuracy, which makes the relative improvement over the Timesformer~\cite{bertasius2021space} and ViViT~\cite{arnab2021vivit} by $1.3\%$ and $0.5\%$ respectively, while it remains more computationally efficient.
This accuracy is also higher than CNN alternatives that either operate in the single RGB modality or use additional flow information. 
Again, it verifies that our model is effective in learning across multiple modalities for the complex video classification task. 

However, we also notice that current video transformers incur relatively large computational overhead compared to several efficient CNN architectures~\cite{lin2019tsm,jiang2019stm,kwon2020motionsqueeze,li2020tea}. 
Therefore, a future line of research could be improving computational and memory efficiency upon the video transformer architectures by leveraging, e.g.,  low-rank approximations~\cite{wang2020linformer}, localized attention span~\cite{liu2021swin} and kernelization~\cite{katharopoulos2020transformers}.
\begin{table}[!thb]
\caption{Performance comparison with state-of-the-art methods on Kinetics600. The inference resolution is denoted by $M\times T\times S^2$ for the number of modalities, temporal and spatial sizes.}
\vglue -0.30cm
\label{tbl:kinetics600}
\centering
\setlength\extrarowheight{1pt}
\resizebox{1\columnwidth}{!}{%
\begin{tabular}{c c c c c c}
\Xhline{1pt}
\multirow{1}{*}{}	
&   \multicolumn{1}{c}{\textbf{Inference}} 
&   \multicolumn{1}{c}{\textbf{Top-1}} 
&   \multicolumn{1}{c}{\textbf{Top-5}} 
&   \multicolumn{1}{c}{\textbf{Inference}} 
\\
\multirow{1}{*}{} 
& \multicolumn{1}{c}{\textbf{Resolution}} 
& 
&
& \multicolumn{1}{c}{\textbf{TFLOPs$\times$views}} 
\\
\hline		
\textbf{CNN +  RGB } & &\\
AttentionNAS~\cite{wang2020attentionnas} & 
$1\times 16 \times 224^2$  &  78.5 & 93.9 & 1.03$\times$1 \\
LGD-3D R101~\cite{qiu2019learning} & 
$1\times 16 \times 224^2$ & 81.5 & 95.6 & -\\
SlowFast R101-NL~\cite{feichtenhofer2019slowfast} & $1\times 16 \times 224^2$ & 81.8 & 95.1 & 0.23$\times$30 \\
X3D-XL~\cite{feichtenhofer2020x3d} & 
$1\times 16 \times 224^2$ & 81.9 & 95.5 & 0.05$\times$30 \\
\hline
\textbf{CNN + RGB + Optical Flow} & & \\
LGD-3D Two-Stream~\cite{qiu2019learning} & $2\times 16\times 112^2$& 83.1 & 96.2 & -\\
\hline
\textbf{Video Transformer} & & \\
Timesformer-L~\cite{bertasius2021space} & 
$1\times 96 \times 224^2$ & 82.2 & 95.6 & 2.38$\times$3\\
ViViT-L~\cite{arnab2021vivit} & 
$1\times 32 \times 320^2$ & 83.0 & 96.5 & 3.99$\times$12\\
MM-ViT & 
$4\times 8 \times 224^2$& 81.5 & 95.1 & 0.82$\times 3$ \\
\textbf{MM-ViT} & 
$4\times 16 \times 224^2$ & \textbf{83.5} & \textbf{96.8} & 1.66 $\times 3$\\
\Xhline{1pt}
\end{tabular}}
\vglue -0.2cm
\end{table}

\subsection{Qualitative Results}
In order to qualitatively evaluate our proposed model, we visualize model attention from the output tokens to the input space via Attention Rollout method~\cite{abnar2020quantifying}.
In Figure.~\ref{fig:attn_vis}, we show examples obtained from applying MM-ViT to UCF-101 videos.
We can see that MM-ViT indeed attends to the relevant regions in the input space.
For example, when applied to classify the video ``Apply Eye Makeup", the model concentrates on the eye area and the eyeshadow brush.
%
%
In some cases, MM-ViT also perceives the phrases/words which are semantically aligned with the content of the action, e.g., the model focuses on words ``lower eyelid" when classifying the video ``Apply Eye Makeup".
The remarkable consistency of the quantitative and qualitative results again confirms the effectiveness of the proposed MM-ViT in complex spatial-temporal-audio reasoning. 

\section{Conclusion}
In this work, we propose \textbf{MM-ViT}, a pure-transformer approach for action recognition in the compressed video domain. 
By factorizing self-attention over the input space, our model enjoys the rich multi-modal information present in the compressed videos, while maintains a reasonable computational cost. 
%
Empirical results also confirm the effectiveness of the proposed cross-modal attention mechanisms in learning the inter-modal relations. 
%
Further, extensive experiments on three public benchmarks show that our model outperforms the competing uni-modal video transformers in both efficiency and accuracy, and performs better or equally well to the state-of-the-art CNN counterparts with the assistance of computationally expensive optical flow. 



{\small
\bibliographystyle{ieee_fullname}
\bibliography{egbib}
}

\end{document}